\documentclass{article} 
\usepackage{iclr2018_workshop,times}
\usepackage[fleqn]{amsmath}
\usepackage{hyperref}
\usepackage{url}
\usepackage{graphicx}
\usepackage{subfigure}
\usepackage{bm}
\usepackage{multirow}
\usepackage{algorithm}
\usepackage{algorithmic}

\title{Stacked Filters Stationary Flow For Hardware-Oriented Acceleration Of Deep Convolutional Neural Networks}

\author{Gao Yuechao, Liu Nianhong \& Zhang Sheng \thanks{ Corresponding author. zhang\_sh@tsinghua.edu.cn } \\
Department of Microelectrics and Nanoelectrics\\
Tsinghua University\\
Beijing, 100084, China \\
\texttt{\{gyc15,lnh15\}@mails.tsinghua.edu.cn}
}

\begin{document}

\maketitle

\begin{abstract}
To address memory and computation resource limitations for hardware-oriented acceleration of deep convolutional neural networks (CNNs), we present a computation flow, stacked filters stationary flow (SFS), and a corresponding data encoding format, relative indexed compressed sparse filter format (CSF), to make the best of data sparsity, and simplify data handling at execution time. And we also propose a three dimensional Single Instruction Multiple Data (3D-SIMD) processor architecture to illustrate how to accelerate deep CNNs by taking advantage of SFS flow and CSF format. Comparing with the state-of-the-art result \citep{han2015deep_compression}, our methods achieve 1.11$\times$ improvement in reducing the storage required by AlexNet, and 1.09$\times$ improvement in reducing the storage required by SqueezeNet, without loss of accuracy on the ImageNet dataset. Moreover, using these approaches, chip area for logics handling irregular sparse data access can be saved (about 19.1\% chip area in \citep{han2016eie}). Comparing with the 2D-SIMD processor structures in DVAS, ENVISION, etc., our methods achieve about 3.65$\times$ processing element (PE) array utilization rate improvement (from 26.4\% to 96.5\%) on the data from Deep Compression on AlexNet\footnote{https://github.com/songhan/Deep-Compression-AlexNet}.
\end{abstract}

\section{Introduction}

CNNs have achieved substantial progress during the past years. But hardware resource limitations have hindered their wide usage in embedded devices. Various efforts have been made to address this issue, such as ShiftCNN \citep{Gudovskiy2017ShiftCNN}, Ristretto \citep{Gysel2016Ristretto}, Eyeriss \citep{Chen2017Eyeriss}, Deep Compression \citep{han2015deep_compression} and EIE \citep{han2016eie}. Through compressing deep neural networks with pruning, trained quantization and Huffman coding, Deep Compression \citep{han2015deep_compression}, the best paper of ICLR 2016, achieved state-of-the-art result in reducing storage requirement of neural networks without affecting their accuracy.

In spite of the great progress achieved till now, there are still many problems to be solved. The first problem is manipulating compressed sparse data need considerable extra logics and consumes extra clock cycles. Eyeriss\citep{Chen2017Eyeriss} uses network on chip (NoC) to handle sparsity by only performing data reads and MACs on nonzero values; DVAS\citep{Moons2015DVAS} and ENVISION\citep{Moons201714} use input guard memories and guard control units to handle data sparsity. Several sparse matrix encoding formats have been proposed, such as CSC, CSR and CISR \citep{Fowers2014A}. But existing encoding formats complex the computation at runtime due to their irregular memory access characteristics. This results in inefficiency in parallelizing computation and bigger chip area. For example, EIE\citep{han2016eie} use Pointer Read Units (accounting for about 19.1\% chip area) and a Sparse Matrix Read Unit (accounting for about 73.57\% chip area) to handle compressed sparse data. Therefore, it would be desirable if the sparse data can be easily handled during execution without complex transformations, lookups and computation. The second one is, for deeply compressed sparse networks, the PE array utilization rate of recently proposed hardware acceleration designs, such as Eyeriss \citep{Chen2017Eyeriss}, DVAS \citep{Moons2015DVAS}, ENVISION \citep{Moons201714}, DNPU \citep{Shin201714}, etc., is fairly low. In this paper we present a novel computation flow SFS, and a corresponding data memory layout and encoding format CSF which achieves the desirable goal that data can be straightforwardly handled at run time. We also propose a three dimensional Single Instruction Multiple Data (3D-SIMD) processor architecture which takes full advantage of these two features.

\section{Stacked filters stationary flow (SFS)}

Computations of convolutional (CONV) and fully connected (FC) layers in CNNs can be unified into one formula Eq.\ref{eq:conv1} (ignoring biases). Eq.\ref{eq:group}-\ref{eq:param} illustrate the approach SFS. ${\bm{V}_{o}}$, ${\bm{V}_{i}}$ and ${\bm{W}_{f}}$ are the matrices of output feature maps, input feature maps and filters, respectively. $S,C,K,M,M',m,W,H,W',H'$ is a given stride size, channel number, filter kernel size, total filter number, number of batches, batch size, input feature width, height and output feature width, height (Eq.\ref{eq:param}). Consider a bank of $M$ filters each with size $K$ and an $H\times W$ feature with $C$ input channels in a layer. We denote the filter bank as a four dimensional array ${\bm{W}_{f}}$ with size $M\times C \times K\times K$, the input feature as a three dimensional array ${\bm{V}_{i}}$ with size $C\times H\times W$, and the output feature as a three dimensional array ${\bm{V}_{o}}$ with size $M\times H'\times W'$. Filters are firstly grouped into $M'$ batches with batch size $m$ (Eq.\ref{eq:group}), and each ${\bm{W}_{f}}^{(n)}$ is then reshaped to ${\bm{W}_{f'}}^{(n)}$ (Eq.\ref{eq:tranwf}). One channel of feature data will convolute with $m$ filters from the same channel in parallel (Eq.\ref{eq:conv2}, $j=0,...,m-1$, pseudo code is illustrated in algorithm \ref{alg:formula}). At the end of computation, ${{\bm{V}_{o'}}^{(0)}}$ - ${{\bm{V}_{o'}}^{(M'-1)}}$ are concatenated back to ${{\bm{V}_{o}}}$ (Eq.\ref{eq:tranvo}).
\begin{align}
&{{\bm{V}_{o}}}[cho][y][x]=\sum\limits_{chi=0}^{C-1}{\sum\limits_{r=0}^{K-1}{\sum\limits_{c=0}^{K-1}{{{\bm{W}_{f}}}[cho][chi][r][c]\times {{\bm{V}_{i}}}[chi][Sy+r][Sx+c]}}} \label{eq:conv1} \\
&{{\bm{W}_{f}}}=[{{\bm{W}_{f}}^{(0)}},{{\bm{W}_{f}}^{(1)}},...,{{\bm{W}_{f}}^{(M'-1)}}], {{\bm{W}_{f'}}}=[{{\bm{W}_{f'}}^{(0)}},{{\bm{W}_{f'}}^{(1)}},...,{{\bm{W}_{f'}}^{(M'-1)}}]\label{eq:group}\\ &{{\bm{W}_{f'}}^{(n)}}[chi][r][c][j]={{\bm{W}_{f}}^{(n)}}[j][chi][r][c]\label{eq:tranwf} \\
&{{\bm{V}_{o'}}^{(n)}}[j][y][x]=\sum\limits_{chi=0}^{C-1}{\sum\limits_{r=0}^{K-1}{\sum\limits_{c=0}^{K-1}{{{\bm{W}_{f'}}^{(n)}}[chi][r][c][j]\times {{\bm{V}_{i}}}[chi][Sy+r][Sx+c]}}} \label{eq:conv2} \\
&{{\bm{V}_{o}}}=[{{\bm{V}_{o'}}^{(0)}},{{\bm{V}_{o'}}^{(1)}},...,{{\bm{V}_{o'}}^{(M'-1)}}]\label{eq:tranvo}
\end{align}
\begin{equation} \label{eq:param} \begin{aligned}
&0\le cho<M, 0\le chi<C, 0\le r<K, 0\le c<K, 0\le j<m,\\
&0\le x<W', 0\le y<H', 0\le n<M', \\
&M'=M/m, W'=(W-K)/S+1, H'=(H-K)/S+1.
\end{aligned} \end{equation}
\begin{algorithm}\scriptsize
\caption{SFS parallel computing pseudo code}
\label{alg:formula}
\begin{algorithmic}
\FOR{each $chi \in [0,C-1]$}
\FOR{each $r \in [0,K-1]$}
\FOR{each $c \in [0,K-1]$}
\STATE $output\ channel\ 0: \ \ \ \ \ \ \ \ \ \ {{\bm{V}_{o'}}^{(n)}}[0][y][x]+={{\bm{W}_{f'}}^{(n)}}[chi][r][c][0]\times {{\bm{V}_{i}}}[chi][Sy+r][Sx+c] $
\STATE $output\ channel\ 1: \ \ \ \ \ \ \ \ \ \ {{\bm{V}_{o'}}^{(n)}}[1][y][x]+={{\bm{W}_{f'}}^{(n)}}[chi][r][c][1]\times {{\bm{V}_{i}}}[chi][Sy+r][Sx+c] $
\STATE $...$
\STATE $output\ channel\ m-1: {{\bm{V}_{o'}}^{(n)}}[m-1][y][x]+={{\bm{W}_{f'}}^{(n)}}[chi][r][c][m-1]\times {{\bm{V}_{i}}}[chi][Sy+r][Sx+c] $
\ENDFOR
\ENDFOR
\ENDFOR
\end{algorithmic}
\end{algorithm}

\section{Relative indexed compressed sparse filter (CSF) format}

\begin{figure}[h]
\begin{center}
\includegraphics[width=12cm]{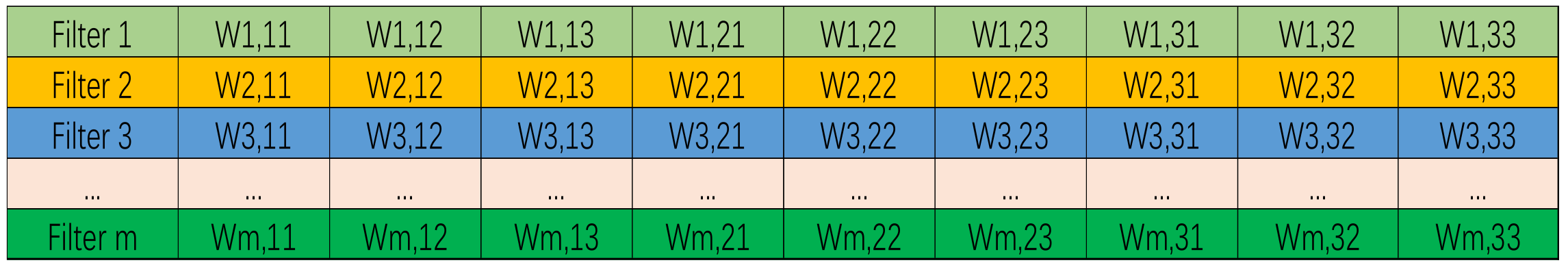}
\end{center}
\caption{Memory layout for the $m$ filters with kernel size 3 from a single channel.}
\label{memory-layout}
\end{figure}

\begin{figure}[h]
\begin{center}
\includegraphics[width=12cm]{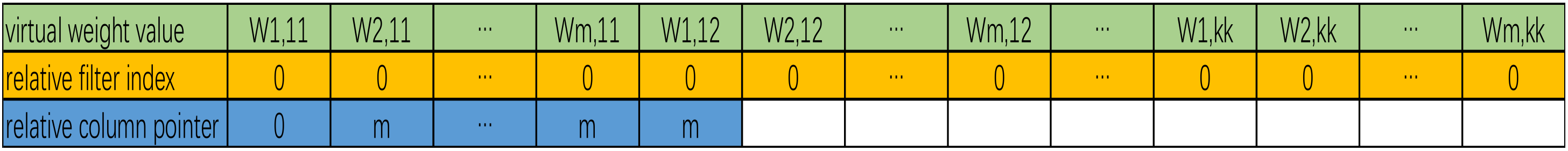}
\end{center}
\caption{Memory layout in the relative indexed CSF format.}
\label{memory-layout-csf}
\end{figure}

As to the encoding format CSF, this approach is to further rearrange the memory layout of the grouped $m$ filters illustrated in figure \ref{memory-layout}, storing the elements column by column. So in computation flow SFS, when each element in the feature map multiplies with a column of data from $m$ filters (algorithm \ref{alg:formula}), the filter weights could be loaded sequentially. The first line in figure \ref{memory-layout-csf} illustrates the changing. In figure \ref{memory-layout-csf}, if there is any weight value equals to 0, just remove that value and its index, add one to the relative index of the next value, and subtract one to the pointer of the next column. The nonzero value number (includes padding zeros) of a column is given by the pointer of the next column. Column pointer is 0 means all the values in the column before the column of this pointer equal to 0. Relative column pointer is not needed when parameters are stored in files.

\section{3D-SIMD processor architecture}
\label{processor}

\begin{figure}[h]
\begin{center}
\includegraphics[width=11cm]{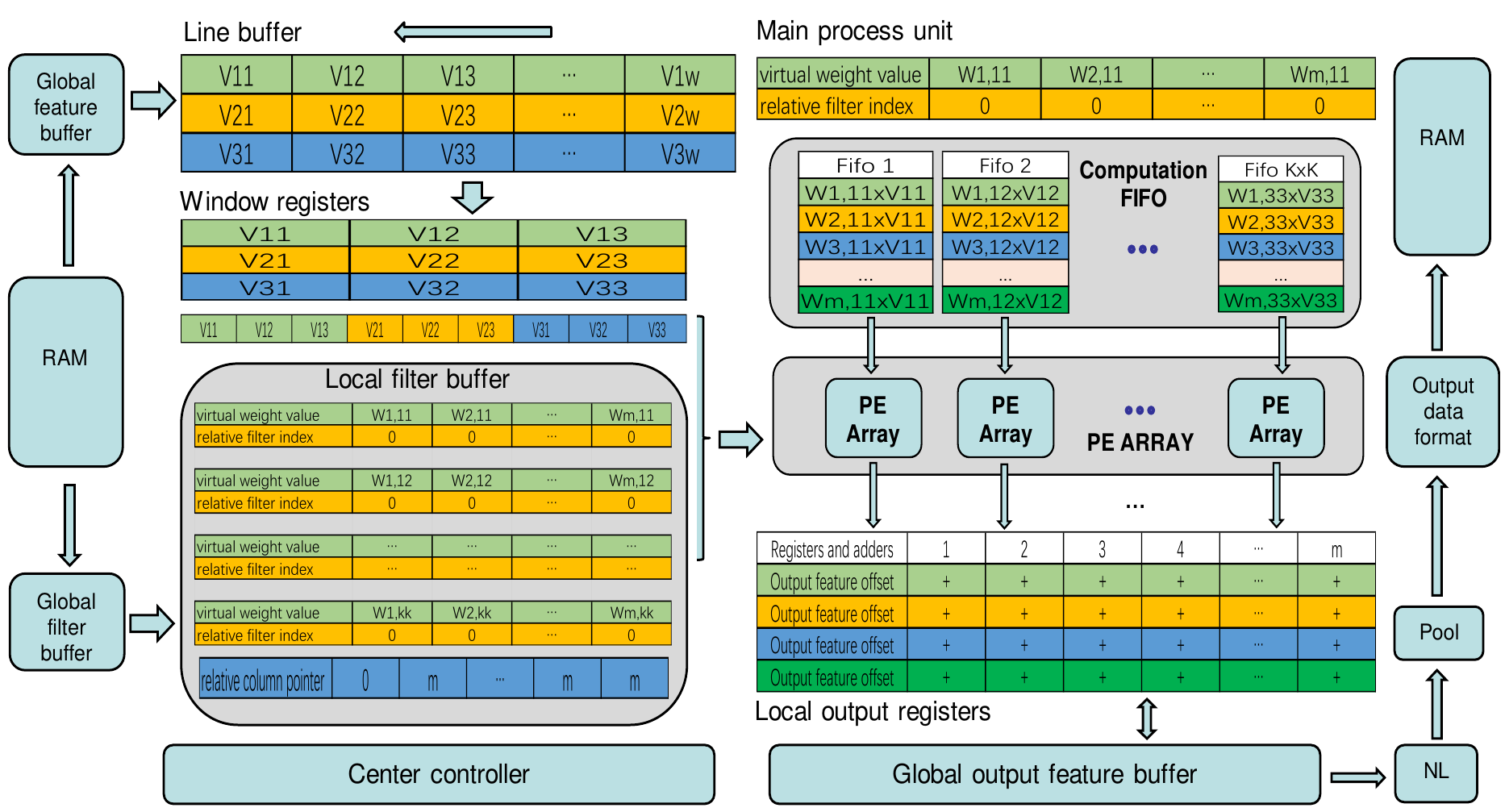}
\end{center}
\caption{3D-SIMD processor architecture.}
\label{3D-SIMD}
\end{figure}

The SFS flow and the CSF encoding format are two key features of the proposed 3D-SIMD processor architecture, see figure \ref{3D-SIMD}. In this architecture, after feature data are loaded into the line buffer and window registers from a single channel of input feature map, and $m$ filter data from the same channel are loaded into the local filter buffer, each element in the window will multiply with a column of data from $m$ filters at the same position (algorithm \ref{alg:formula}). So data in CSF format can be straightforwardly handled without complex transformations, lookups and computation and loaded sequentially at run time, and zeros are skipped as designed.  This demonstration shows that these two approaches can greatly simplify sparse data handling, saving zero bypassing and data lookup time. There are no complex sparse data handling logics needed comparing with former works\citep{Moons2015DVAS, Moons201714, han2016eie}.

\section{Result}

\begin{figure}[h]
\begin{center}
\includegraphics[width=14cm]{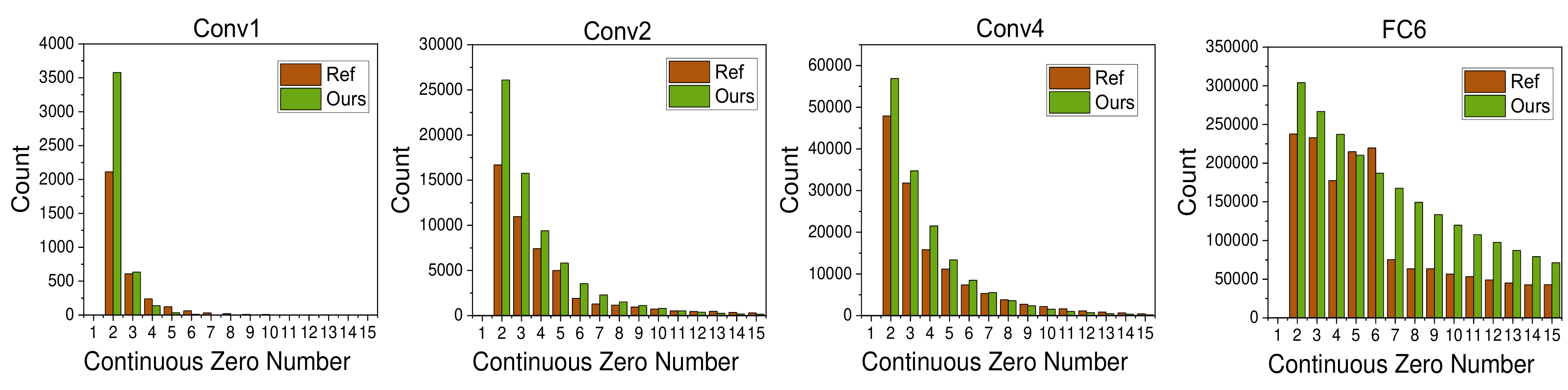}
\end{center}
\caption{Distributions of continuous zero numbers in different layers of Alexnet, comparing with reference work \citep{han2015deep_compression}.}
\label{zero-distribution}
\end{figure}

The distribution of continuous zero numbers after applying the changing is first evaluated. As figure \ref{zero-distribution} shows, the distribution narrows to the left. It means that fewer bits are needed to store the relative index values, and there will be fewer padding zeros when compressing data in encoding formats. This will further reduce storage space. The distribution of continuous nonzero numbers after applying the changes is also evaluated. The distribution also narrows to the left, which means the computation load during execution will be better balanced comparing to the reference work\citep{han2015deep_compression}. The effect of batch size $m$ on storage space is also analyzed. It shows that there do exist an optimum batch size for each layer. Smaller batch size requires smaller local buffer, but data are less reused and the input feature map need to be loaded more times. For simplicity, all the experiments in this section use filter number as the batch size. That means, all the filters and input feature maps are loaded only one time and the output feature maps are saved one time for one reference. Eq.\ref{eq:best_bits} is used to find the best bit number to store the relative index values of each layer. Table~\ref{extra-space-table} and table~\ref{compression-table} illustrate the improvement of extra space needed to store index and padding zeros in each layer of Alexnet comparing to former works and the improvement of total storage requirement after applying our method. The PE array utilization rate\footnote{PE array utilization rate is estimated by: no. of nonzero value MACs / total no. of MACs.} improvement of convolutional and fully connected layers on several networks are also evaluated. On Alexnet, as illustrated in table~\ref{utilization-table}, the number of MACs of dense network is 1.06GOPS, the number of MACs of nonzero weight value of sparse network is 0.28GOPS, and the number of MACs after applying our method is 0.29GOPS. So comparing with dense network processor like the 2D-SIMD processor structures in DVAS, ENVISION, etc., the PE array utilization rate improves from 26.4\% to 96.5\% (about 3.65$\times$ improvement), using the data from Deep Compression on AlexNet. The amount of data lookup calculation\footnote{Single calculation of locating a batch of data is defined as a basic unit.} is also evaluated. Using the same data above, the amount of calculation of SFS and CSF approach is about $1/20$ that of the algorithm in EIE (see table~\ref{utilization-table}).
\begin{equation}
\label{eq:best_bits}
\underset{bit}{\mathop{\arg \min }}\,\{{{f}_{total\_bits}}(bit)=Nz\_num\times bit+\sum\limits_{i={{2}^{bit}}}^{{{\max }}}{\text{zero}\_stat[i]\times (\frac{i}{{{2}^{bit}}})}\times (wbit+bit)\}
\end{equation}
$Nz\_num$: the number of nonzero weight values;
$wbit$: the number of bits to store weight value;
$bit$: the number of bits to store relative index value;
$zero\_stat$: the distribution of continuous zero numbers.

\begin{table}[hp]\footnotesize
\caption{Space(in bits) needed in each layer of Alexnet for storing extra padding zeros and index}
\label{extra-space-table}
\begin{center}
\begin{tabular}{lllll}
\multicolumn{1}{c}{\bf Layer}  &\multicolumn{1}{c}{\bf Nonzeros} &\multicolumn{1}{c}{\bf Index(bit)} &\multicolumn{1}{c}{\bf Extra space} &\multicolumn{1}{c}{\bf Improvement}\\
\hline
conv1 in \citep{han2015deep_compression} &235088 &4 &117568 & \\
conv1 by SFS+CSF &235088 &1 &37189 &3.16$\times$\\
conv2 in \citep{han2015deep_compression} &930448 &4 &491456 & \\
conv2 by SFS+CSF &930448 &3 &393897 &1.25$\times$ \\
conv3 in \citep{han2015deep_compression} &2447520 &4 &1262136 & \\
conv3 by SFS+CSF &2447520 &3 &1082226 &1.17$\times$ \\
conv4 in \citep{han2015deep_compression} &1975696 &4 &999260 & \\
conv4 by SFS+CSF &1975696 &3 &822902 &1.21$\times$\\
conv5 in \citep{han2015deep_compression} &1304496 &4 &662352 & \\
conv5 by SFS+CSF &1304496 &3 &545220 &1.21$\times$\\
\hline
fc6 in \citep{han2015deep_compression} &13345544 &4 &23978248\\
fc6 by SFS+CSF &13345544 &5 &18849823 &1.27$\times$\\
fc7 in \citep{han2015deep_compression} &6149136 &4 &9525904 & \\
fc7 by SFS+CSF &6149136 &5 &8373849 &1.14$\times$\\
fc8 in \citep{han2015deep_compression} &4204128 &4 &4289032 & \\
fc8 by SFS+CSF &4204128 &3 &4033864 &1.06$\times$\\
\hline
total in \citep{han2015deep_compression} &30592056 & &41325956\\
total by SFS+CSF &30592056 & &34138970 &1.21$\times$\\
\end{tabular}
\end{center}
\end{table}

\begin{table}[hp]\footnotesize
\caption{Extra space(in bits) improvement and total storage requirement improvement}
\label{compression-table}
\begin{center}
\begin{tabular}{lllll}
\multicolumn{1}{c}{\bf Network}  &\multicolumn{1}{c}{\bf Nonzeros} &\multicolumn{1}{c}{\bf Extra space} &\multicolumn{1}{c}{\bf Improvement }&\multicolumn{1}{c}{\bf Total}\\
\hline
AlexNet by \citep{han2015deep_compression}              &30592056 &41325956 \\
AlexNet by SFS+CSF            &30592056 &34138970 &1.21$\times$ &1.11$\times$ \\
\hline
SqueezeNet by \citep{han2015deep_compression}              &3327368 &1737628 \\
SqueezeNet by SFS+CSF        &3327368    &1307160 &1.33$\times$ &1.09$\times$
\end{tabular}
\end{center}
\end{table}

\begin{table}[hp]\footnotesize
\caption{PE array utilization rate and data lookup calculation improvement (MACs in GOPS)}
\label{utilization-table}
\begin{center}
\begin{tabular}{llllll}
\multirow{2}{1cm}{\bf }  &\multirow{2}{1.4cm}{\bf Total no. of MACs} &\multirow{2}{2.2cm}{\bf No. of nonzero value MACs} &\multirow{2}{1.8cm}{\bf Total no. of MACs (CSF)}&\multirow{2}{1.3cm}{\bf Speed-up (CSF)} &\multirow{2}{1.3cm}{\bf Lookup (CSF)}\\
\\
\hline
Alexnet CONV layers   &1.00269368	&0.2744399	&0.2839878	&3.53$\times$ & 1/13\\
Alexnet FC layers    &0.05459595	&0.0055178	&0.0059305	&9.21$\times$  & 1/42\\
Alexnet CONV+FC layers  &1.05728963	&0.2799577	&0.2899182	&3.65$\times$ & 1/20\\
\hline
PE untilization ratio        &0.2647881	&	&0.9656438	
\end{tabular}
\end{center}
\end{table}

\section{Conclusion}

In this paper, we propose a stacked filters stationary flow SFS, and its corresponding data encoding format CSF. And we also propose a 3D-SIMD processor architecture for this computation flow and data encoding format. Experimental results show that our approaches narrow the distribution of the numbers of continuous zeros and nonzeros to the smaller number direction. This helps to further compress the network parameters by about 8\% to 10\% and balance computation load at run time. By adopting the proposed 3D-SIMD processor architecture, chip area for logics handling irregular memory access of sparse data can be saved, for example, about 19.1\% chip area in EIE \citep{han2016eie} for pointer read can be saved. Moreover, directly using the encoded data without complex transformations, lookups and computation at runtime can also save zero bypassing clock cycles, and data lookup time \citep{Moons2015DVAS, Moons201714, han2016eie}.

\bibliography{iclr2018_workshop}
\bibliographystyle{iclr2018_workshop}

\end{document}